%% file: main.tex
\documentclass[10pt,twocolumn,letterpaper]{article}

\usepackage{cvpr}              
\usepackage{multirow}

\input{preamble}

\definecolor{cvprblue}{rgb}{0.21,0.49,0.74}
\usepackage[pagebackref,breaklinks,colorlinks,citecolor=cvprblue]{hyperref}

\def\paperID{5836} 
\def\confName{CVPR}
\def\confYear{2024}

\title{Generalized Deepfakes Detection with Reconstructed-Blended Images and Multi-scale Feature Reconstruction Network}

\author{Yuyang Sun$^{1,2}$, Huy H. Nguyen$^2$, Chun-Shien Lu$^3$, ZhiYong Zhang$^4$, Lu Sun$^5$ and Isao Echizen$^{1,2}$ \\
\small{$^1$The University of Tokyo, Japan\ \ \ \ \ \ \ \ \ \ 
$^2$National Institute of Informatics, Japan\ \ \ \ \ \ \ \ \ \ 
$^3$Academia Sinica, Taiwan} \\
\small{$^4$Sun Yat-sen University, China\ \ \ \ \ \ \ \ \ \ 
$^5$South China University of Technology, China} \\
{\tt\small \{tarrysun, nhhuy,iechizen\}@nii.ac.jp}
}

\begin{document}
\maketitle
\input{sec/0_abstract}    
\input{sec/1_intro}
\input{sec/2_related}
\input{sec/3_method}
\input{sec/4_experiment}
\input{sec/5_conclusion}
\section*{Acknowledgments}
This work was partially supported by JSPS KAKENHI Grant JP21H04907, and by JST CREST Grants JPMJCR18A6 and JPMJCR20D3, Japan.
{
    \small
    \bibliographystyle{ieeenat_fullname}
    \bibliography{main}
}
\input{sec/X_suppl}


\end{document}

%% file: preamble.tex
\usepackage[dvipsnames]{xcolor}


%% file: sec/0_abstract.tex
\begin{abstract}
The growing diversity of digital face manipulation techniques has led to an urgent need for a universal and robust detection technology to mitigate the risks posed by malicious forgeries. We present a blended-based detection approach that has robust applicability to unseen datasets. It combines a method for generating synthetic training samples, i.e., reconstructed blended images, that incorporate potential deepfake generator artifacts and a detection model, a multi-scale feature reconstruction network, for capturing the generic boundary artifacts and noise distribution anomalies brought about by digital face manipulations. Experiments demonstrated that this approach results in better performance in both cross-manipulation detection and cross-dataset detection on unseen data. 
\end{abstract}

%% file: sec/1_intro.tex
\section{Introduction}
\label{sec:intro}

Digital facial manipulation, exemplified by deepfake technology, entails substituting or altering facial attributes, expressions, and appearances to create highly deceptive visual content. The advancement of deep generative models \cite{goodfellow2014generative,karras2017progressive,karras2019style,karras2020analyzing,choi2018stargan,choi2020stargan} and computer vision technologies has facilitated the creation of efficient, automated deepfake pipelines (e.g., \cite{FaceApp}), enabling non-experts to easily manipulate visual media content. However, this ease of use has led to increasing misuse in various domains (politics, journalism, mass media, etc.) and violations of individual privacy.

Early detection models \cite{yang2019exposing,agarwal2019protecting,afchar2018mesonet,nguyen2019capsule,li2018exposing,sabir2019recurrent,sun2021improving} are susceptible to overfitting specific artifacts and noise patterns. They are deficient in cross-domain detection, resulting in substantial performance deterioration when confronted with unseen datasets. Researchers thus introduced blended-based deepfake detection methods \cite{li2020face,shiohara2022detecting,zhao2021learning} that create blended samples through the fusion of suitable genuine facial samples, thereby simulating prevalent manipulation artifacts such as misalignment, incongruent boundary textures, and irregularities in pixel or color distribution.

\begin{figure}[t]
	\begin{center}
		\includegraphics[width=1\linewidth]{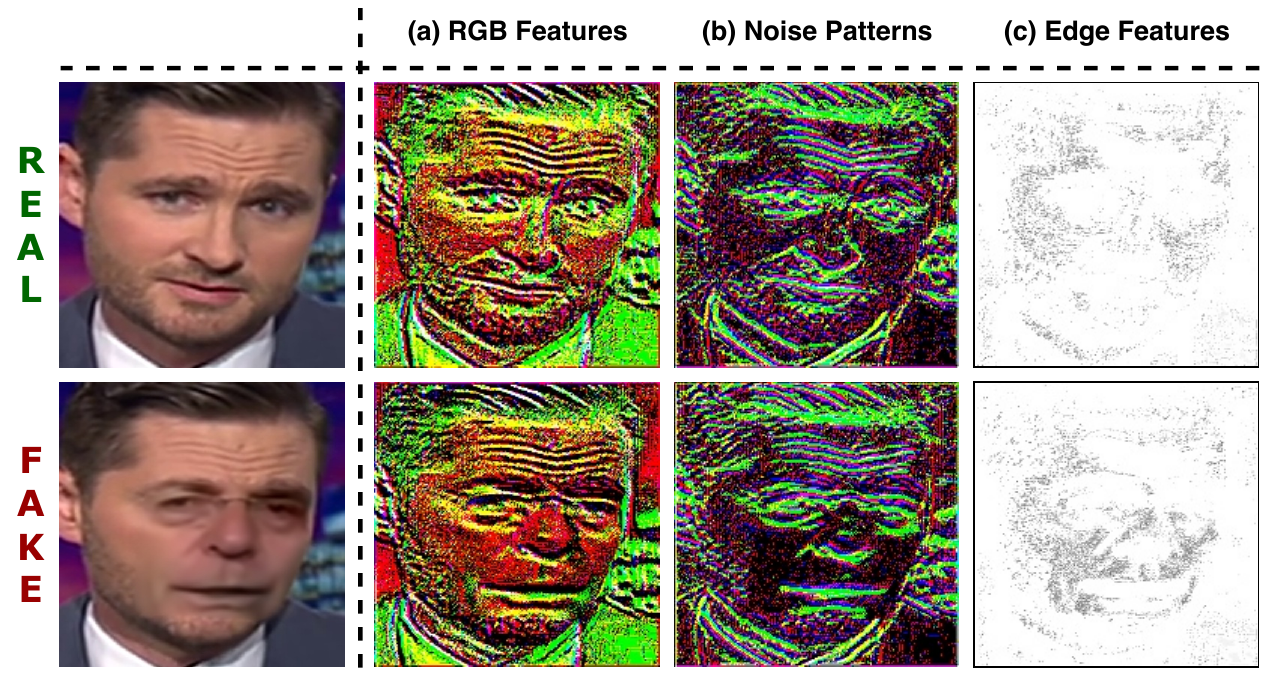}
	\end{center}
	\caption{(a) RGB features, (b) noise patterns, and (c) edge features extracted from the first block of EfficientNet-B4 for real image (top row) and corresponding fake image (bottom row). }
	\label{fig:fig1}
\end{figure}

Previous methods for constructing blended samples typically extract facial entities from the original foreground after they undergo certain augmentations and then graft them onto alternative original backgrounds. This results in synthetic blended samples that lack certain invisible artifacts, including unique generator fingerprints \cite{mccloskey2018detecting,koopman2018detection} and anomalies in channel distribution introduced by adaptive instance normalization \cite{huang2017arbitrary}. Furthermore, previous methods for detecting forgeries predominantly perform binary classification with foundational backbones or learned self-supervised consistency features when training from blended samples, resulting in inadequate harnessing of the potential of diverse annotations inherent in synthesized samples.

We have devised a novel approach to detecting forgeries that combines a method for generating synthetic forgery training samples and a detection model for capturing the generic boundary artifacts and noise distribution anomalies brought about by digital face manipulation. Our reconstructed blended image (RBI) method builds upon advanced deepfake generation techniques \cite{li2019faceshifter,chen2020simswap,wang2021hififace,xu2022styleswap,kim2022smooth} to disentangle identity and background information from genuine facial images. Random Gaussian noise is infused into a background vector with a predefined probability, and the identity-background pairs are passed through a decoder to obtain reconstructed images. This process incorporates latent generator patterns and distinctive fingerprints into the synthesized blended data. After application of the statistical transformations of Shiohara and Yamasaki \cite{shiohara2022detecting}, the reconstructed image is blended with the original image and a mask is used to generate a training sample. 

Our detection model, a multi-scale feature reconstruction network (MFRN), effectively exploits diverse training artifacts and manipulated regions. Built upon prior research\cite{li2020face,shiohara2022detecting,das2022gca,chen2021image}, it focuses on the three possible origins of forged features shown in Figure \ref{fig:fig1}: RGB features, noise patterns, and edge features. RGB features include color inconsistencies between inner and outer facial regions and an abnormal image channel covariance, noise patterns include mismatched image frequencies in manipulated regions and invisible generator fingerprints, and edge features include boundary conflicts arising from face splicing and landmark misalignment. A specifically designed architecture is used to extract the corresponding features from the input and combine them at multiple scales to self-supervise the reconstruction of blended boundaries and regions.

We conducted cross-manipulation and cross-dataset detection assessments on unseen data to align with real-world detection scenarios. Our model was trained solely on genuine data from the FF++ dataset \cite{rossler2018faceforensics}. For cross-manipulation detection, it achieved areas under the curve (AUCs) of 100\%, 100\%, 99.88\%, 99.81\%, and 98.90\% respectively, for DeepFake (DF) \cite{Deepfakes}, Face2Face (F2F) \cite{thies2016face2face}, FaceSwap (FS) \cite{FaceSwap}, NeuralTextures (NT) \cite{thies2019deferred}, and FaceShifter (FSH) \cite{li2019faceshifter} manipulation. For cross-dataset detection, it demonstrated robust detection on prominent deepfake detection datasets such as CDF-v2 \cite{li2020celeb} (AUC of 95.27\%), DFD \cite{DFD} (99.12\%), DFDC \cite{dolhansky2020deepfake} (73.31\%), and DFDC-P \cite{dolhansky2019deepfake} ((83.66\%). These results surpass or match those of current state-of-the-art baselines. 

Our contributions are summarized as follows:

\begin{itemize}
	\item We introduce a method for generating simulated forgery training samples (reconstructed-blended images) that enhances the diversity of simulated artifacts by incorporating an invisible generator fingerprint and noise pattern.
	\item We introduce a model (a multi-scale feature reconstruction network) for harnessing the diversity in training artifacts and manipulated regions present in blended samples.
	\item We validated the performance of this approach across multiple unseen manipulation techniques and datasets and demonstrated its superiority.
\end{itemize}

%% file: sec/2_related.tex
\section{Related Work}
\label{sec:related}

\subsection{Deepfake Generation}
Modern deepfake generation techniques primarily use deep generative models \cite{goodfellow2014generative, kingma2013auto,ho2020denoising,rombach2022high} to achieve seamless integration and migration of facial features and expressions, enabling creation of highly realistic visual content. Earlier deepfake models \cite{Deepfakes,FaceSwap} separated the process of generating and blending forged faces, leading to noticeable stitching artifacts and easy human eye detection.

Subsequent deepfake models addressed this problem by combining the generation and blending into a single step. Incorporating the concept of style transfer, these models consider the target identity information as a style feature and use techniques like AdaIN \cite{huang2017arbitrary} to align the target identity with the source background, resulting in end-to-end deepfake generation pipelines that improve realism and reduce visible artifacts. Several of these models \cite{li2019faceshifter,wang2021hififace,xu2022styleswap} predict masks for the target background, enabling precise control over manipulated areas. Others \cite{chen2020simswap,kim2022smooth} do not require such masks and instead autonomously learn the content that needs to be modified or retained during training.

\begin{figure*}
	\begin{center}
		\includegraphics[width=0.87\linewidth]{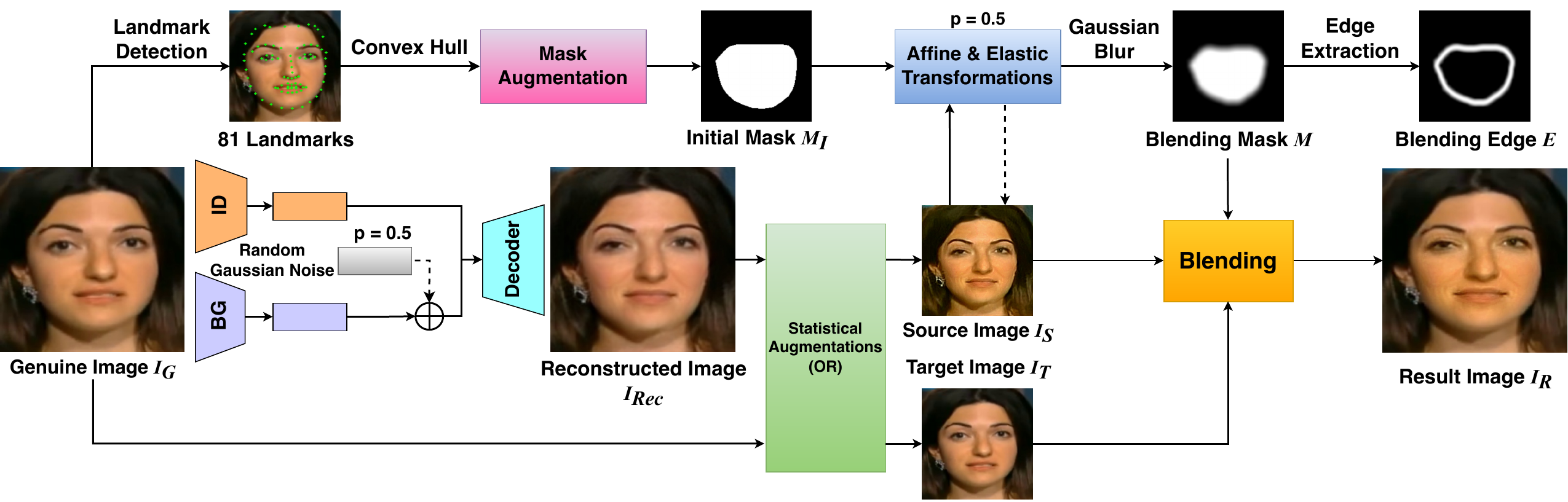}
	\end{center}
	\caption{\textbf{Illustration of RBI generation.} Information in given genuine image $I_G$ is disentangled, and image is reconstructed as $I_{Rec}$, incorporating a unique fingerprint and noise patterns. Statistical transformations are then randomly applied to create source image $I_S$ and target image $I_T$ in which common visible manipulation artifacts are simulated. Concurrently, $I_G$'s convex hull is augmented and deformed to produce mask $M$ and edge $E$, which are used guide the blending of $I_S$ and $I_T$, yielding result image $I_R$.}
	\label{fig:fig2}
\end{figure*}

\subsection{Deepfake Detection}
Detection techniques that combine spatial-frequency statistical analysis with deep neural networks have shown marked efficacy in identifying image manipulation artifacts. These methods \cite{chen2021image,sun2022edge,das2022gca,guo2023hierarchical} focus on detecting abnormal edges and noise features introduced by splicing together disparate source images, resulting in the capture of imperceptible manipulation artifacts. Deepfake detection, a specialized domain within image manipulation detection, can be categorized into artifact-based and blended-based detection.

\textbf{Artifact-based Detection} is aimed at identifying potential deepfakes by detecting prevalent and distributed artifacts or imperfections introduced during manipulation. Early research focused on visible artifacts such as abnormal facial expressions and head movements \cite{agarwal2019protecting}, inconsistencies in predicted head pose \cite{yang2019exposing}, image photo response non-uniformity \cite{koopman2018detection}, and color cues \cite{mccloskey2018detecting}. As forged images have become increasingly realistic, manually specified artifacts have proven insufficient. Subsequent research \cite{nguyen2019capsule,afchar2018mesonet,luo2021generalizing,zhao2021multi,qian2020thinking} has focused on using neural networks to detect spatial-frequency pattern differences within manipulated images. Additionally, researchers have observed that frame-by-frame manipulation of videos can introduce temporal inconsistencies at the interframe level, leading to efforts \cite{zheng2021exploring,sabir2019recurrent,guera2018deepfake,masi2020two,sun2023face} to uncover the absence of temporal correlations within video streams.

\textbf{Blended-based Detection} is aimed at identifying potential deepfakes by using forgery training samples generated by blending pairs of genuine samples that have undergone flexible image augmentations through diverse blending, creating an array of manipulation artifacts. These methods aim for high generalization performance on unseen datasets by incorporating various manipulation patterns.  Early attempts in this category, such as DSP-FWA (Dual Spatial Pyramid for Exposing Face Warp Artifacts in DeepFake Videos) \cite{li2018exposing}, simulate distortion artifacts introduced by affine transformations in the deepfake pipeline by blending downsampled and Gaussian-blurred facial regions with the background. Subsequent methods like Face X-Ray \cite{li2020face} and patch-wise consistency learning (PCL)+ inconsistency image generation (I2G) \cite{zhao2021learning} simulate blending boundaries or inconsistent information by replacing and blending similar facial images from the dataset. The use of self-blended \cite{shiohara2022detecting} images has improved detection performance and achieved higher data efficiency by generating pseudo source-target image pairs through statistical transformations applied to a single image, followed by blending.

\begin{figure*}
	\begin{center}
		\includegraphics[width=1\linewidth]{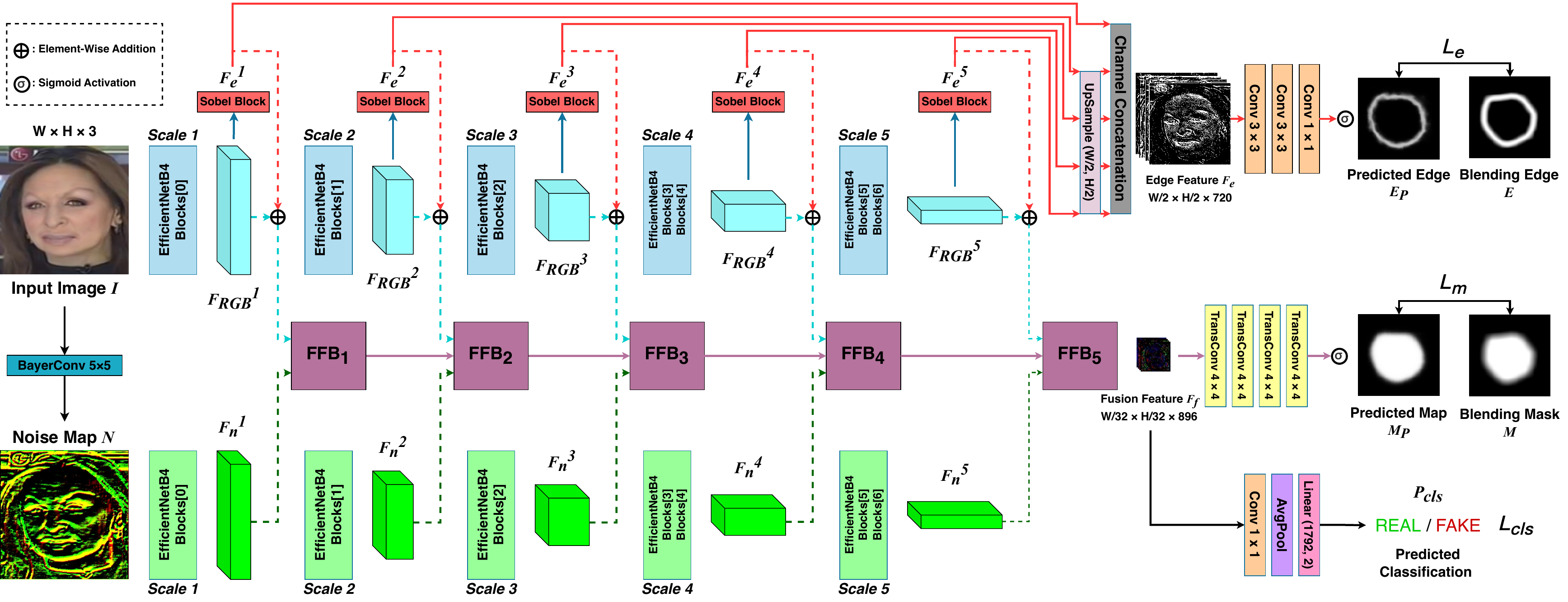}
	\end{center}
	\caption{\textbf{Overview of MFRN architecture.} First, noise map $N$ of input image $I$ is acquired. Next, RGB features $F_{RGB}$ and noise patterns $F_n$ are independently extracted across diverse scales by using separate backbones. $F_{RGB}$ undergoes Sobel block to create corresponding edge features $F_{e}$ at their respective scales, which are upsampled for manipulation boundary prediction ($E_P$). Concurrently, a feature fusion block integrates features from the diverse scales for manipulation map prediction ($M_p$) and authenticity classification.}
	\label{fig:fig3}
\end{figure*}

%% file: sec/3_method.tex
\section{Proposed Approach}

In our approach to enhancing the generalization of deepfake detection, we introduce a novel synthetic training sample, namely the reconstructed blended image (RBI), and a corresponding detection framework, namely multi-scale feature reconstruction network (MFRN).

As illustrated in Figure \ref{fig:fig2}, RBI is generated by extracting identity and background features from the source image, embedding imperceptible latent generator artifacts, integration of common visible image artifacts through statistical transformations, and finally blending with the source image through a randomly deformed mask. This results in diversity and enrichment in the synthetic training samples. Some examples of RBIs and detailed explanations of relative artifacts can be found in Figure \ref{fig:fig6} and Section \ref{sec:rationale} in the appendix.

Once the samples have been generated alongside the corresponding blending masks and boundary annotations, as depicted in Figure \ref{fig:fig3}, a convolutional neural network-based reconstruction model, the MFRN, is created. This model captures RGB, edge, and noise features across various scales, thus enabling comprehension of the mapping from blended samples to region labels. This enables the model to autonomously identify discrepancies introduced by manipulation and to detect conflicting boundary textures.

\subsection{Reconstructed Blended Image Generation}

Genuine image $I_{G}$ is disentangled using an identity (ID) encoder and background (BG) encoder. The resulting identity-background pair is then decoded, producing reconstructed image $I_{Rec}$. This step disrupts innate noise patterns within the genuine image and thereby introduces a simulated generator fingerprint. Encoding is diversified by adding random Gaussian noise with mean ($\mu$) 0 and standard deviation ($\sigma$) in $[0.1, 0.3]$ to the background vector with probability $p=0.5$. Note that this encoder-decoder framework can be replaced with an analogous deepfake generator that combines identity and background features. We used a pretrained SimSwap \cite{chen2020simswap} model to perform the reconstruction.

Next, in accordance with the method outlined in \cite{shiohara2022detecting}, 
statistical augmentations are equally applied to either $I_{G}$ or $I_{Rec}$, forming source ($I_{S}$) and target ($I_{T}$) images for blending. This integration introduces evident spatial or frequency irregularities like those commonly encountered in manipulated images. Specifically, following operations are executed on the each image: random RGB channel value shift, random adjustment of hue, saturation, and brightness, random modification of brightness and contrast, and random application of blurring or sharpening. 

Concurrently, 81 landmarks are extracted from $I_{G}$ using the Dlib library \cite{king2009dlib} and are used to form initial mask $M_{I}$ by convex hull calculation. The hull is then diversified using the $random\_get\_hull$ from open source tool \cite{mask} for mask augmentation. With a probability of $p=0.5$, identical affine and elastic transformations are applied to both $M_{I}$ and $I_{S}$, simulating landmark discrepancies and boundary conflicts introduced by warping in the deepfake pipeline.

Finally, transformed $M_{I}$ is subjected to Gaussian blur, yielding blending mask $M$, which is used to guide the blending process. The blending of $I_{S}$ and $I_{T}$ using $M$ to obtain $I_{R}$ is done using

\begin{equation}
I_R = I_S \odot \alpha M + I_T \odot (1 - \alpha M) 
\end{equation}
where $\alpha \in [0.5,1]$ regulates the magnitude of the blending mask. Corresponding blending edge $E$ can be readily extracted \cite{li2020face}:
\begin{equation}
E = 4 \cdot M \odot (1 - M) 
\end{equation}

\subsection{Multi-scale Feature Reconstruction Network}
\begin{figure}[t]
	\begin{center}
		\includegraphics[width=0.9\linewidth]{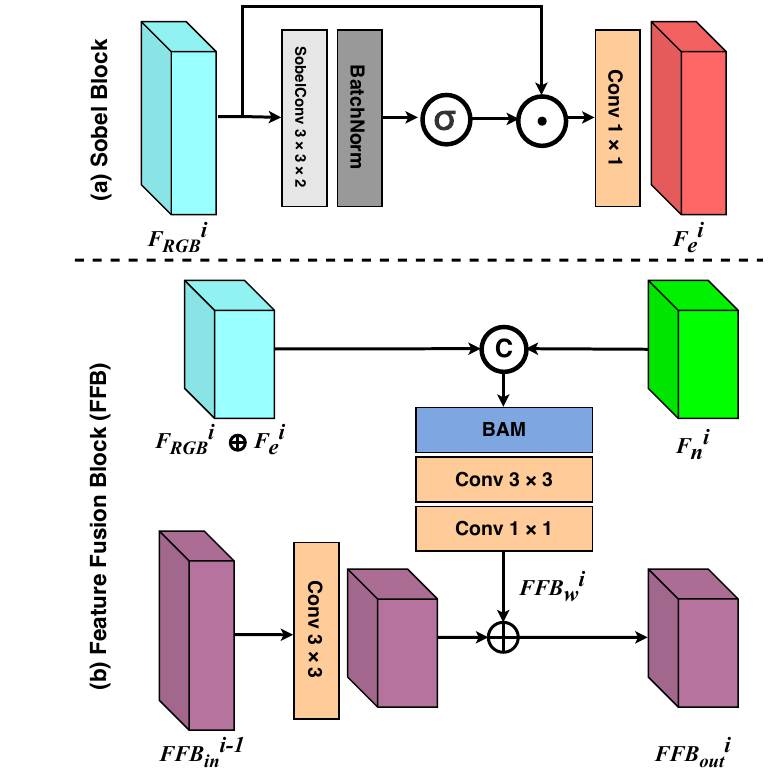}
	\end{center}
	\caption{Architecture of Sobel block and feature fusion block.}
	\label{fig:fig4}
\end{figure}

Input image $I \in \mathbb{R}^{W \times H \times 3}$ is converted to grayscale, and noise map $N$ is derived using a $5 \times 5$ Bayer constrained convolutional layer \cite{bayar2018constrained}. Two separate EfficientNet-B4 \cite{tan2019efficientnet} backbones are then used to capture features from both $I$ and $N$. For clarity, the output of the $i^{th}$ downsampling block is designated as being at scale $i$. At each scale $i$, feature maps $F_{RGB}^i$ and $F_n^i$ are extracted from the RGB and noise branch, respectively. The edge features are computed from the RGB feature using the Sobel block depicted in Figure \ref{fig:fig4}(a), which involves applying two fixed-parameter $3 \times 3$ Sobel filters to $F_{RGB}^i$, followed by batch normalization \cite{ioffe2015batch} and Sigmoid activation. The resulting output is element-wise multiplied with $F_{RGB}^i$, and integration is achieved using a $1 \times 1$ convolutional layer to produce $F_e^i$. This procedure can be summarized by the following formula:
\begin{equation*}
Edge\_Act_i = \sigma((BN(SobelConv(F_{RGB}^i)))) 
\end{equation*}
\begin{equation}
F_e^i = Conv(F_{RGB}^i \odot Edge\_Act_i)   
\end{equation}

Subsequently, the extracted feature maps are split into two pathways and used to individually reconstruct the modified boundaries and regions within the blended image.

For one pathway, $F_e^i$ at each scale is uniformly upsampled to $(\frac{W}{2}, \frac{H}{2})$, which, after channel-wise concatenation, yields feature $F_e \in \mathbb{R}^{\frac{W}{2} \times \frac{H}{2} \times 720}$. Two $3 \times 3$ convolutional kernels are used to capture local information, and a $1 \times 1$ convolutional kernel is used to integrate channel features, finally producing predicted edge $E_p \in \mathbb{R}^{\frac{W}{2} \times \frac{H}{2}}$.

For another pathway, feature fusion block (FFB) is introduced to facilitate both within-scale feature fusion and across-scale feature propagation. As illustrated in \ref{fig:fig4}(b), at scale $i$, FFB$_{i}$ receives $F_n^{i}$, the element-wise summation of $F_{RGB}^{i}$ and $F_e^{i}$, and the output feature from FFB$_{i-1}$ (excluding the first FFB block) as input. The rationale behind summing $F_{RGB}^{i}$ and $F_e^{i}$ stems from their shared provenance in the RGB domain. Subsequently, a bottleneck attention module (BAM) \cite{park2018bam} applied to the concatenated extracted features establishes a self-attention mechanism that directs attention to anomalies within manipulated areas on the basis of the spatial disposition and channel distribution of feature map. Following the processing of the two convolutional layers, weighted feature $FFB_{w}^i$ is obtained for the present scale. A $3 \times 3$ convolutional kernel is used to align the size of the propagated feature originating at the previous scale, denoted as $FFB_{in}^{i-1}$, with that of the one at the current scale. Subsequent element-wise summation with the current weighted features results in the output feature $FFB_{out}^i$, which is then propagated to FFB$_{i+1}$ after ReLU activation. These operations can be succinctly encapsulated:

\begin{equation*}
FFB_{w}^i = Conv_{\times 2}(BAM(Cat(F_{RGB}^{i} \oplus F_e^{i}, F_n^{i}))))
\end{equation*}
\begin{equation}
FFB_{out}^i = FFB_{w}^i \oplus Conv(FFB_{in}^{i-1})
\end{equation}

After the features of the different modalities at the various scales are fused and passed hierarchically, fusion feature $F_f \in \mathbb{R}^{\frac{W}{32} \times \frac{H}{32} \times 896}$ is obtained. Several transposed convolutions with a kernel size of 4 are used to predict modification map $M_p \in \mathbb{R}^{\frac{W}{2} \times \frac{H}{2}}$. Furthermore, a parallel classification branch, encompassing a convolutional head, a pooling layer, and a dense layer, is created and used to predict two-dimensional classification vector $P_{cls}$, which is derived from the $F_f$ feature.

Clearly, the lack of manipulation in authentic instances means that outcomes $E_p$ and $M_p$ for a genuine image should manifest as matrices consisting exclusively of zeros.

\subsection{Loss Functions}
\textbf{Map Loss and Edge Loss.} Element-wise binary cross-entropy (BCE) loss is used to assess the dissimilarity between the predicted map and the blending mask, as well as between the predicted edge and the blending edge:
\begin{equation}
\small
L_{e} = -\frac{1}{N} \sum_{i,j}(E^{i,j}logE_p^{i,j} + (1-E^{i,j})log(1-E_p^{i,j}))
\end{equation}
\begin{equation}
\small
L_{m} = -\frac{1}{N} \sum_{i,j}(M^{i,j}logM_p^{i,j} + (1-M^{i,j})log(1-M_p^{i,j}))
\end{equation}
where $N=\frac{WH}{4}$ is the number of pixels in the $E_p$ and $M_p$.

\noindent \textbf{Classification Loss.} BCE loss is used to quantify the classification error, given by
\begin{equation}
L_{cls} =T_{cls}logP_{cls} + (1-T_{cls})log(1-P_{cls})
\end{equation}
where $T_{cls}=\{0,1\}$ represents the ground truth label of the input sample.

The overall loss function of our model is
\begin{equation}
L = \lambda_1L_{m} + \lambda_2L_{e} + L_{cls}
\end{equation}
where $\lambda_1$ and $\lambda_2$ are scaling factors used to regulate individual loss proportions.

%% file: sec/4_experiment.tex
\section{Experiments}
\subsection{Setup}

\begin{table*}[t]
\centering
\small
\renewcommand\arraystretch{1.18}
\resizebox{0.75\textwidth}{!}{%
\begin{tabular}{lccccccc}
\toprule[1.5pt] 
\multirow{2}{*}{\textbf{Method}} & \multicolumn{7}{c}{Test AUC (\%)}                                                                                  \\ \cline{2-8} 
                        & DF           & F2F            & FS             & NT             & FF++(w/o FSH)  & FSH            & FF++            \\ \hline
Face X-Ray \cite{li2020face}              & 99.17        & 98.57          & 98.21          & 98.13          & 98.52          & -              & -              \\
PCL + I2G \cite{zhao2021learning}              & \textbf{100} & 98.97          & 99.86          & 97.63          & 99.11          & -              & -              \\
Self-blended \cite{shiohara2022detecting}            & 99.99        & 99.88          & \textbf{99.91} & 98.79          & 99.64          & -              & -              \\
Self-blended \textbf{*} \cite{shiohara2022detecting}          & 99.94        & 99.72          & 99.76          & 98.23          & 99.42          & 97.70          & 99.07          \\ \hline
RBIs + MFRN (Ours)      & \textbf{100} & \textbf{100} &    99.88       & \textbf{99.81} & \textbf{99.92} & \textbf{98.90} & \textbf{99.71} \\ \bottomrule[1.5pt] 
\end{tabular}}
\caption{\textbf{Cross-manipulation comparison among blended-based methods.} Results are cited from the papers, with \textbf{*} indicating official pre-trained model results on our test data. Our method outperformed baselines on DF, F2F, NT, FSH, and the complete FF++ dataset.}
\label{tab:tab2}
\end{table*}

\noindent \textbf{Training Data.} Our model was trained solely on pristine FaceForensics++ \cite{rossler2018faceforensics} (FF++) data, following the official dataset split, which included 720 training videos. We uniformly extracted 20 frames from each video and utilized Dlib \cite{king2009dlib} to extract 81 facial landmarks from each frame, which were used to compute the initial mask. We used the RetinaFace face detector \cite{deng2020retinaface} for bounding box localization and face cropping. In cases with multiple faces detected in a video frame, we overlapped the bounding boxes with a corresponding deepfake manipulation mask, selecting the result with the largest intersection area for validity.

\noindent \textbf{Test Data.} We assessed model performance through cross-manipulation and cross-dataset evaluation. For cross-manipulation evaluation, we used unseen manipulation videos from the FF++ test set, including Deepfakes \cite{Deepfakes} (DF), Face2Face \cite{thies2016face2face} (F2F), FaceSwap \cite{FaceSwap} (FS), NeuralTextures \cite{thies2019deferred} (NT), and FaceShifter \cite{li2019faceshifter} (FSH), totaling 840 test videos (140 from the pristine dataset and each manipulation dataset), following official dataset splits. For cross-dataset evaluation, we utilized several prominent digital face manipulation datasets: DeepFakeDetection \cite{DFD} (DFD), Celeb-DF-v2 \cite{li2020celeb} (CDF-v2), DeepFake Detection Challenge \cite{dolhansky2020deepfake} (DFDC), and its preview version \cite{dolhansky2019deepfake} (DFDC-P). We used the complete DFD dataset for testing and adhered to official data splits for the other datasets, using their designated test videos. For all videos, we uniformly sampled frames and used RetinaFace for bounding box extraction and face cropping. As the current models had reached saturation in FF++ detection performance, we assessed cross-manipulation at the individual sample level, extracting and cropping five faces per video to construct the test set. Conversely, we evaluated cross-dataset performance at the video level, extracting 32 frames evenly, cropping faces, and aggregating model predictions for all 32 faces per video on the basis of their mean. To ensure fairness, videos where face extraction was not possible were assigned a prediction value of 0.5.

\begin{table}[t]
\centering
\normalsize
\renewcommand\arraystretch{1.35}
\resizebox{0.49\textwidth}{!}{%
\begin{tabular}{lccccc}
\toprule[2pt] 
\multirow{2}{*}{\textbf{Method}} & \multirow{2}{*}{Type}           & \multicolumn{4}{c}{Test AUC (\%)}                                 \\ \cline{3-6} 
                                 &                                 & CDF-v2         & DFD            & DFDC           & DFDC-P         \\ \hline
DeepRhythm \cite{qi2020deeprhythm}                       & \multirow{5}{*}{Artifact} & -              & -              & -              & 64.1           \\
MultiATT \cite{zhao2021multi}                        &                                 & 67.44          & -              & -              & -              \\
FRDM \cite{luo2021generalizing}                            &                                 & 79.4           & 91.9           & -              & 79.7           \\
RECCE \cite{cao2022end}                           &                                 & 68.71          & -              & 69.06          & -              \\
FTCN \cite{zheng2021exploring}                            &                                 & 86.9           & -              & -              & 74.0           \\ \hline
Face X-Ray \cite{li2020face}                      & \multirow{3}{*}{Blended}  & -              & 93.47          & -              & 71.15          \\
PCL + I2G \cite{zhao2021learning}                       &                                 & 90.03          & \underline{99.07}    & 67.52          & 74.37          \\
Self-blended \cite{shiohara2022detecting}                    &                                 & \underline{93.18}    & 97.56          & \underline{72.42}    & \textbf{86.15} \\ \hline
RBIs + MFRN (Ours)               & Blended                   & \textbf{95.27} & \textbf{99.12} & \textbf{73.31} &  \underline{83.66}    \\ \bottomrule[2pt] 
\end{tabular}}
\caption{\textbf{Cross-dataset detection comparison with baselines.} With best and second-best results indicated in bold and underline. Our method outperformed the baselines on CDF-v2, DFD, and DFDC and was second best on DFDC-P.}
\label{tab:tab1}
\end{table}

\noindent \textbf{Comparison Baseline.} We used advanced baselines to evaluate our model, categorizing them into two types: artifact-based and blended-based. The former type requires both genuine and manipulated samples for training and included Multi-Attentional Deepfake Detection \cite{zhao2021multi} (MultiAtt), Fusion + RSA + DCMA + Multi-scale \cite{luo2021generalizing} (FRDM), Uncovering Common Feature \cite{yan2023ucf} (UCF), Spatial-Phase Shallow Learning \cite{liu2021spatial} (SPSL), Reconstruction-Classification Learning \cite{cao2022end} (RECCE), Fully Temporal Convolution Network \cite{zheng2021exploring} (FTCN) and DeepRhythm \cite{qi2020deeprhythm}. In contrast, the latter type requires only genuine samples and included Face X-Ray \cite{li2020face}, PCL+I2G \cite{zhao2021learning}, and Self-blended\cite{shiohara2022detecting}.

\noindent \textbf{Implementation Details.} To generate RBIs, we used the official pre-trained SimSwap model as the reconstruction generator and applied common image augmentations, such as JPEG compression, brightness-contrast adjustments, and color jittering, to both synthesized RBIs and genuine samples. The MFRN used the pre-trained EfficientNet-b4 \cite{tan2019efficientnet} backbone. All facial regions cropped by bounding boxes were resized to $380 \times 380$ to match pre-training specifications. We set $\lambda_1$ and $\lambda_2$ to 100 and 50, respectively; additional insights into the effect of varying loss weights $\lambda$ are available in Table \ref{tab:tab7} in the appendix. To enhance stability and generalization performance, we used the sharpness-aware minimization (SAM) optimizer \cite{foret2020sharpness}. Training was performed over 80 epochs on a NVIDIA A100 (80G) GPU, with a learning rate of 0.001 and a batch size of 32.

\noindent \textbf{Evaluation Metrics.} Due to the highly imbalanced distribution of genuine and manipulated samples in the datasets used for evaluation, we primarily used the AUC as our evaluation metric as it better reflects the model's performance.

\subsection{Cross-Manipulation Detection}
We conducted cross-manipulation detection experiments on the FF++ raw data to evaluate our model's detection ability on unseen manipulations. We maintained fairness by comparing our model with similar blended-based models, citing their reported results directly. Additionally, for assessing the performance of FSH, we utilized the official pre-trained Self-blended model on our test data (the results for Face X-Ray and I2G+PCL could not be replicated due to the lack of official implementations), indicated by an asterisk (\textbf{*}).

As shown in Table \ref{tab:tab2}, in scenarios nearing performance saturation, our model achieved AUCs of 100\%, 100\%, and 99.81\% for common DF, F2F, and NT manipulations, respectively, surpassing the baseline models. It slightly under-performed with a 98.88\% result on FS. It outperformed Self-blended by 1.2\% on FSH, illustrating its enhanced ability to capture generator fingerprints and noise patterns introduced by one-stage deepfake generators. Our method consistently achieved optimal detection AUCs on the complete FF++ test set, whether considering FSH or not, with scores of 99.71\% and 99.92\%, respectively.

\subsection{Cross-Dataset Detection}
As previously mentioned, we validated our method's performance across multiple unseen mainstream forgery datasets. We compared our method's performance with those of artifact-based and blended-based detection methods.

As evident in Table \ref{tab:tab1}, blended-based methods have a pronounced advantage over artifact-based ones when used in cross-dataset detection tasks. This observation supports our statement above that models trained on limited manipulated data are susceptible to overfitting specific artifacts and noise patterns, making them deficient in cross-domain detection. Among the artifact-based methodologies, FTCN, which takes into consideration interframe temporal features, demonstrated superior generalization capabilities, achieving an AUC of 86.9\% on CDF-v2. Among the blended-based methods, ours had better outcomes. Its performance surpassed that of the compared baselines with AUCs of 95.27\%, 99.12\%, and 73.31\% on CDF-v2, DFD, and DFDC, respectively. On DFDC-P, its AUC of 83.66\% is slightly lower than that of Self-blended (86.15\%) and better than those of the other compared baselines. In summation, our model consistently delivered superior performance in cross-dataset detection scenarios.

To assess our method's wider applicability, we also conducted training on samples from varying dataset (CDF-v2) and varying compression level (FF-c23), followed by cross-dataset performance evaluation on unseen datasets in a consistent manner. For the latter, to ensure fairness, we selected three top-performing methods from the DeepfakeBench \cite{yan2023deepfakebench}, which were trained on FF-c23, as comparative baselines. As shown in Table \ref{tab:tab6}, our model shows robust adaptability to unseen data, demonstrating its efficacy even when trained on different dataset or low-quality samples. From the results trained on FF-23, it outperformed prominent state-of-the-art baselines, emphasizing its advantages and affirming its position as a robust solution in the field.

\begin{table}[t]
\centering
\normalsize
\renewcommand\arraystretch{1.35}
\resizebox{0.49\textwidth}{!}{%
\begin{tabular}{lccccc}
\toprule[1.6pt] 
\multirow{2}{*}{\textbf{Method}} & \multirow{2}{*}{\textbf{Training Set}} & \multicolumn{4}{c}{Test AUC (\%)} \\ \cline{3-6} 
                                  &                                        & CDF-v2  & DFD    & DFDC  & DFDC-P \\ \hline
Self-blended \cite{shiohara2022detecting}                      & \multirow{2}{*}{CDF-v2}                & 93.74   & -      & -     & 81.10  \\
RBIs + MFRN (Ours)                &                                        & 93.53   & 98.25  & 73.40 & 85.21  \\ \hline
SPSL \cite{liu2021spatial}                             & \multirow{4}{*}{FF-c23}                & 76.50   & 81.22  & 70.40 & 74.08  \\
UCF \cite{yan2023ucf}                              &                                        & 75.27   & 80.74  & 71.91 & 75.94  \\
FRDM \cite{luo2021generalizing}                              &                                        & 75.52   & 81.20  & 69.95 & 74.08  \\
RBIs + MFRN (Ours)                &                                        & \textbf{93.89}   & \textbf{98.39}  & \textbf{72.70}  & \textbf{81.72}  \\ \bottomrule[1.6pt] 
\end{tabular}}
\caption{Cross-dataset performance of model trained on the CDF-v2 dataset and FF-c23 data. The results of SPSL, UCF, and FRDM were excerpted from DeepfakeBench \cite{yan2023deepfakebench}.}
\label{tab:tab6}
\end{table}

\subsection{Video Compression Robustness}
In the context of digital face manipulation detection, a model's resilience to compression is pivotal given that real-world detection scenarios often entail acquiring highly compressed samples from the Internet. We conducted evaluations using test samples from videos subjected to moderate (c23) and heavy (c40) compression that were taken from the FF++ dataset. Performance was compared with those of Face X-Ray and Self-blended. The latter's results were derived by assessing the pretrained model on our test data.

As shown in Table \ref{tab:tab3}, Face X-Ray is vulnerable to compression, with AUCs of 87.35\% and 61.60\% on c23 and c40, respectively. Our model demonstrates robust detection performance under moderate compression conditions but falls short in heavily compressed scenarios when compared with the simpler detection structure. This discrepancy may stem from our noise pattern extraction and RGB feature specification, which are affected by the substantial loss of image information in heavily compressed data.

\begin{table}[t]
\centering
\large
\renewcommand\arraystretch{1.24}
\resizebox{0.48\textwidth}{!}{%
\begin{tabular}{lccc}
\toprule[1.7pt] 
\multirow{2}{*}{\textbf{Method}} & \multicolumn{3}{c}{AUC on Different Video Compressions (\%)} \\ \cline{2-4} 
                         & \ \ c0 (raw)                & \ \ \ \ \ \qquad  c23               & \ \qquad c40               \\ \hline
Face X-Ray \cite{li2020face}              & \ \ 98.52              & \ \ \ \ \ \qquad 87.35             & \ \qquad 61.60             \\
Self-blended \textbf{*} \cite{shiohara2022detecting}             & \ \ 99.42         & \ \ \ \ \ \qquad 88.32            & \ \qquad 65.29          \\ \hline
RBIs + MFRN (Ours)       & \ \ 99.92              & \ \ \ \ \ \qquad 88.55              & \ \qquad 64.63                 \\ \bottomrule[1.7pt] 
\end{tabular}}
\caption{Robustness for different levels of compression of FF++ dataset, with \textbf{*} indicating official pre-trained model results on our prepared test data. }
\label{tab:tab3}
\end{table}

\subsection{Ablation Study}
\noindent \textbf{Effectiveness of RBIs + MFRN method.} We conducted two intermediate experiments, denoted as self-blended images “SBIs + MFRN” and “RBIs + EfficientNet-b4” to assess the effectiveness of our proposed method. As shown in Table \ref{tab:tab4}, MFRN enhanced the identification of manipulated artifacts, surpassing the performance of the original EfficientNet-b4 backbone and yielding superior detection performance. The inclusion of RBIs introduces diverse visible artifacts and invisible noise anomalies, enhancing the authenticity and diversity of the synthetic training data when compared with the use of SBIs and thereby also contributes to improved detection performance.

\noindent \textbf{Effectiveness of Components in MFRN.} We conducted experiments with the following settings to validate the effectiveness of each component in MFRN: 1) Edge reconstruction branch removed, and RGB features and noise patterns fused at multiple scales to reconstruct manipulated regions;
2) FFB removed, and, when reconstructing manipulated regions, edge features, RGB features, and noise patterns at last scale are concatenated and used as inputs;
3) Both FFB and noise pattern branch removed, and, when reconstructing manipulated regions, edge features and RGB features at last scale are concatenated and used as inputs.

\begin{figure*}
	\begin{center}
		\includegraphics[width=0.92\linewidth]{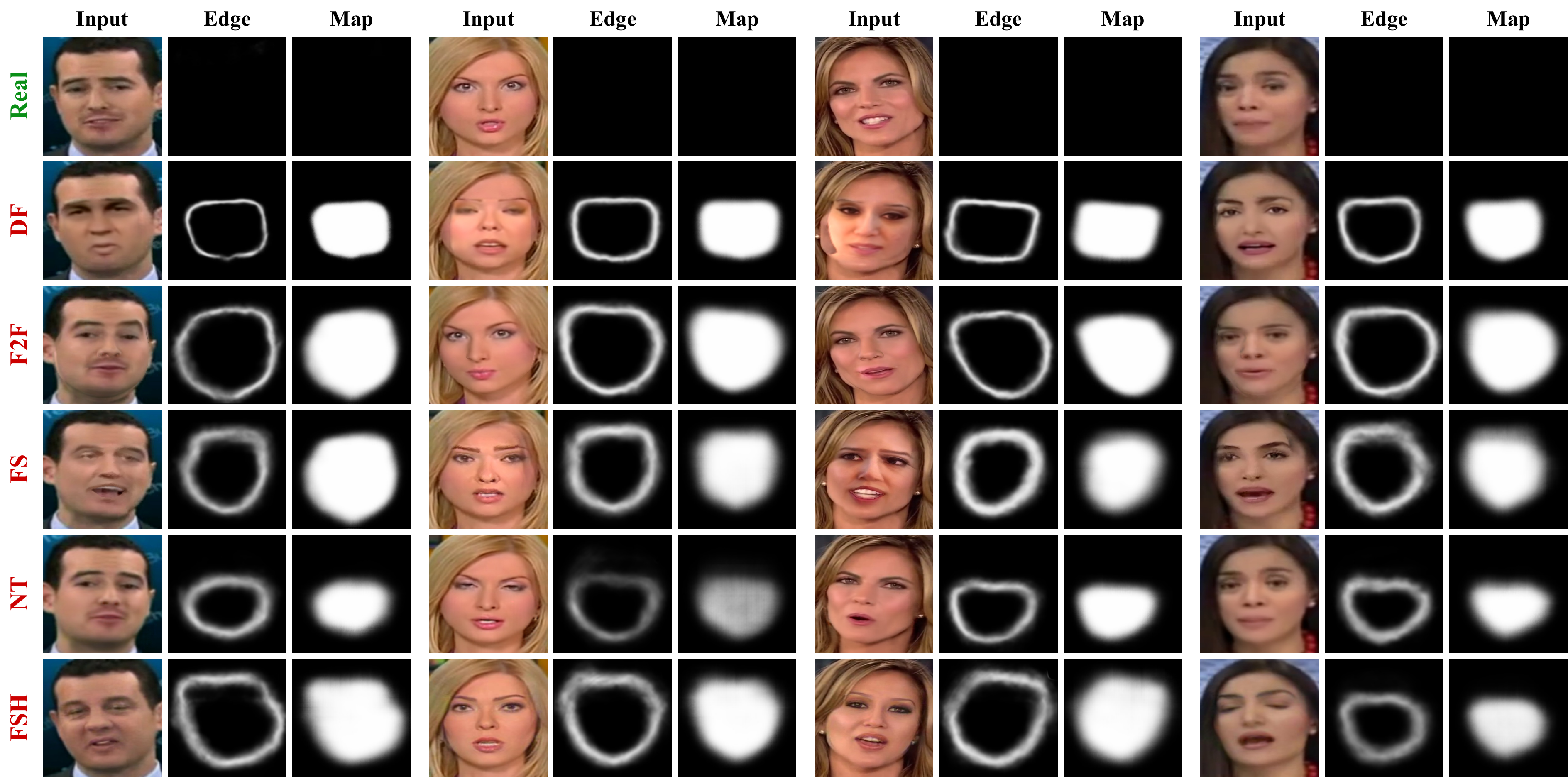}
	\end{center}
	\caption{Visualization of predicted edges and maps for genuine images alongside various manipulated samples. Model was trained on real data from FF++ dataset augmented by RBIs. Inputs used for prediction came from unseen FF++ test dataset.}
	\label{fig:fig5}
\end{figure*}

As shown in Table \ref{tab:tab5}, the absence of the edge reconstruction branch resulted in substantially poorer detection performance. This was due to manipulation techniques often causing unnatural boundary conflicts in the replaced areas, and edge reconstruction effectively aids the model in capturing these misaligned artifacts. In complex detection backgrounds like DFDC-P, models without the noise branch exhibited a notable decrease in detection performance. Furthermore, our FFB effectively boosted the model's detection performance through multi-scale self-attention-based fusion of diverse feature modalities.

\begin{table}[t]
\centering
\scriptsize
\renewcommand\arraystretch{1.3}
\resizebox{0.47\textwidth}{!}{%
\begin{tabular}{lcccc}
\toprule[1.3pt]
\multirow{2}{*}{\textbf{Settings}} & \multicolumn{3}{c}{Test AUC (\%)}                & \multirow{2}{*}{Avg.} \\ \cline{2-4}
                          & CDF-v2         & DFD            & FF++           &                      \\ \hline
SBIs + EfficientNet-b4    & 93.18          & 97.56          & 99.64          & 
96.79                \\
RBIs + EfficientNet-b4       & 92.51          & 98.64        & 99.72          & 96.96                \\ 
SBIs + MFRN               & 95.21          & 98.27      & 99.83          & 97.77                \\ \hline
RBIs + MFRN (Ours)        & \textbf{95.27} &      \textbf{99.12}         & \textbf{99.92} & \textbf{98.10}       \\ \bottomrule[1.3pt]
\end{tabular}}
\caption{Results of the ablation study on RBIs and MFRN.} 
\label{tab:tab4}
\end{table}

\subsection{Visualization}

We visualized prediction outcomes on real and manipulated samples from the unseen FF++ test set. These inputs, used with a model trained exclusively on FF++ real data augmented by RBIs, produced prediction manipulation regions and boundary conflicts, as seen in Figure \ref{fig:fig5}. Our approach effectively aids the model in recognizing irregular patterns in unseen manipulated data and accurately delineating the boundaries of replaced regions, even without direct training on similar manipulations. This effectiveness can be attributed to the use of comprehensive synthetic training data and our purpose-designed feature reconstruction network. More visualization results and explanations can be found in Figure \ref{fig:fig7} and Section \ref{sec:visu} in the appendix.

\begin{table}[t]
\small
\renewcommand\arraystretch{1.25}
\resizebox{0.47\textwidth}{!}{%
\begin{tabular}{lcccc}
\toprule[1.5pt]
\multirow{2}{*}{\textbf{Settings}} & \multicolumn{3}{c}{Test AUC (\%)}                                  & \multirow{2}{*}{Avg.} \\ \cline{2-4}
                          & CDF-v2               & DFDC-P                  & FF++                 &                      \\ \hline
w/o  Edge Reconstruction  &  93.34                    &  82.71               &  99.86                    & 91.97                  \\                  
w/o FFB  & 94.26                & 83.25             & \textbf{99.94}  &     92.48 \\ 
w/o Noise Pattern + FFB         &  95.02              & 79.45            & 99.90                &  91.46   \\ \hline
RBIs + MFRN (Ours)        & \textbf{95.27}                & \textbf{83.66}              & 99.92       & \textbf{92.95}               \\ \bottomrule[1.5pt]
\end{tabular}}
\caption{Results of the ablation study to evaluate the effectiveness of individual components within MFRN.} 
\label{tab:tab5}
\end{table}

%% file: sec/5_conclusion.tex
\section{Conclusion}
We have presented an innovative method for synthesizing forgery training samples, i.e., reconstructed blended images (RBIs). It improves the ability to simulate manipulation artifacts by seamlessly integrating simulated generator fingerprints and noise patterns. We also presented a novel detection model, the multi-scale feature reconstruction network (MFRN), which adeptly exploits the richness of diversity introduced by the use of random blending masks and boundaries within the RBIs. Experimental results demonstrated that our proposed approach substantially enhances performance in both unseen cross-dataset and cross-manipulation detection.

%% file: sec/X_suppl.tex
\clearpage
\setcounter{page}{1}
\maketitlesupplementary

\begin{figure*}[b]
	\begin{center}
		\includegraphics[width=0.96\linewidth]{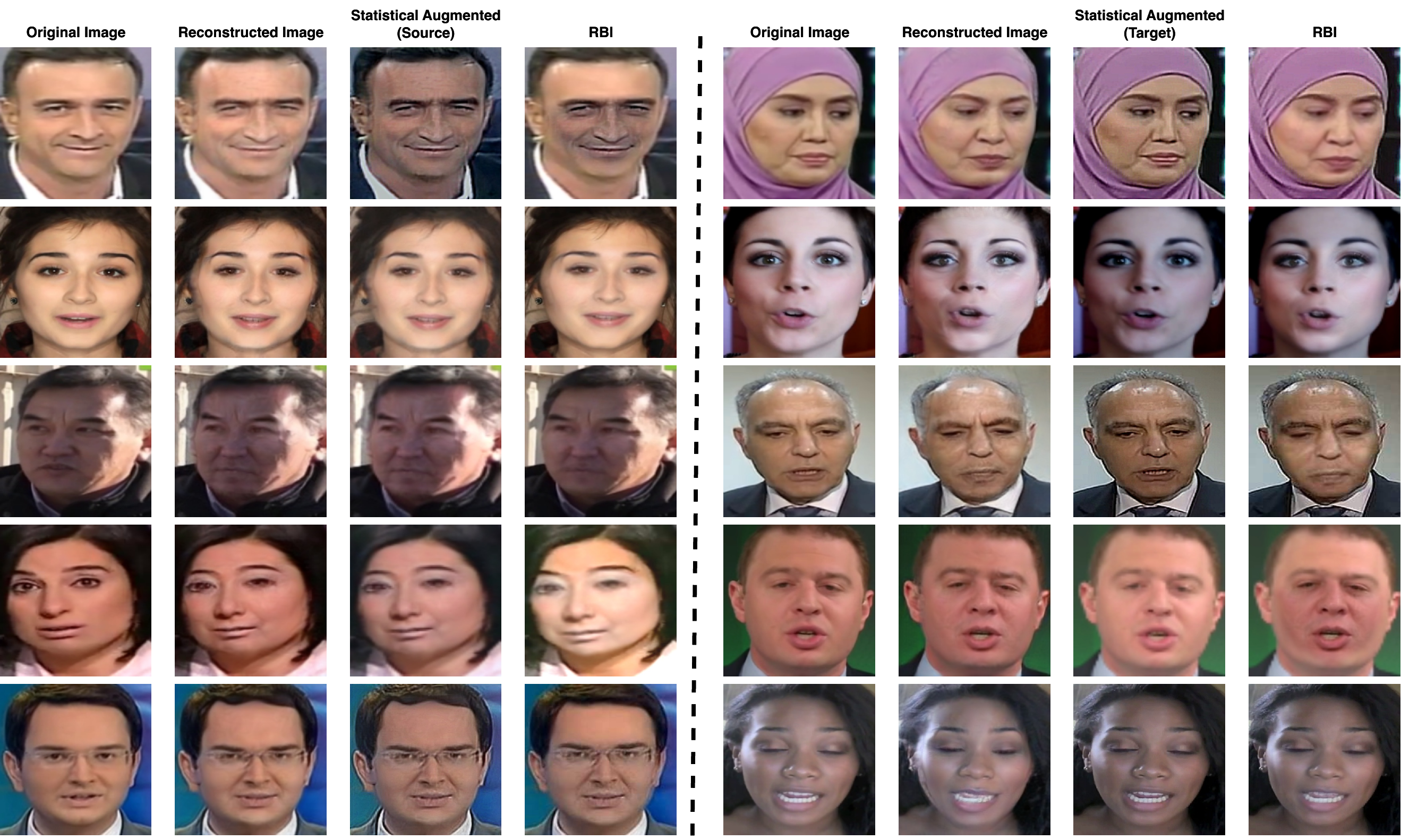}
	\end{center}
	\caption{Examples of RBIs, where the left column represents statistical augmentations applied to the source image (i.e., reconstructed image), and the right column represents statistical augmentations applied to the target image (i.e., original image).}
	\label{fig:fig6}
\end{figure*}

\section{Effect of $\lambda$ in loss function}
We experimentally evaluated the effect of scaling factors $\lambda_1$ and $\lambda_2$ in the loss function on model performance. Keeping the other experimental settings constant, we used several sets of values for hyper-parameters $\lambda_1$ and $\lambda_2$. The results on CDF-v2, DFDC-P, and FF++ are shown in Table \ref{tab:tab7}.

From the results in the table, it can be observed that the hyperparameter settings with $\lambda_1=50$ and $\lambda_2=100$ achieve the best performance.

\section{Samples of RBIs}
\label{sec:rationale}

Several of the RBI samples we generated are shown in Figure \ref{fig:fig6}, with the left half showing the results of statistical augmentation on the foreground face taken from the source image (i.e., the reconstructed image) and with the right half showing the results of statistical augmentation on the background face taken from the target image (i.e., the original image). 

The reconstructed images reveal that our proposed disentanglement-reconstruction process not only introduces visually imperceptible frequency noise as a generator fingerprint (left half, rows 1, 3, and 5; right half, row 4) but also introduces unique visible artifacts created by the generator which cannot be simulated by statistical augmentation. These artifacts include abrupt cheek contours (left half, row 2, right half, row 5), inconsistent eye sizes (right half, row 2), overlapping eyes (left half, row 4), and blurred teeth artifacts (right half, row 5). Introducing pattern noise and distinctive generator artifacts can thus help the model learn more robust and generalizable forgery features, thereby improving the model's detection performance on unseen manipulations and deepfake data.

\begin{table}[]
\centering
\small
\renewcommand\arraystretch{1.15}
\resizebox{0.46\textwidth}{!}{%
\begin{tabular}{cccccc}
\toprule[1.6pt]
\multicolumn{2}{c}{\textbf{Settings}} & \multicolumn{4}{c}{Test AUC (\%)} \\ \cline{3-6} 
$\lambda_1$      & $\lambda_2$       & CDF-v2  & DFDC-P   & FF++   & Avg.  \\ \hline
25           & 25            & 92.84   & 77.52  & 99.83  & 90.06 \\
25           & 50            & 93.96   & 80.51  & 99.89  & 91.45 \\
50           & 50            & 94.51   & 83.85  & 99.81  & 92.72 \\
50           & 100           & 95.27   & 83.66  & 99.92  & 92.95 \\
100          & 50            & 95.17   & 81.78  & 99.91  & 92.29 \\
150          & 300           & 94.28   & 74.50  & 99.87  & 89.55 \\
500          & 1000          & 92.90   & 74.89  & 99.05  & 88.95 \\ \bottomrule[1.6pt]
\end{tabular}}
\caption{Effect of $\lambda$ in loss function on model performance.}
\label{tab:tab7}
\end{table}

\section{More Visulization Results of Our Model}
\label{sec:visu}
\begin{figure*}
	\begin{center}
		\includegraphics[width=1\linewidth]{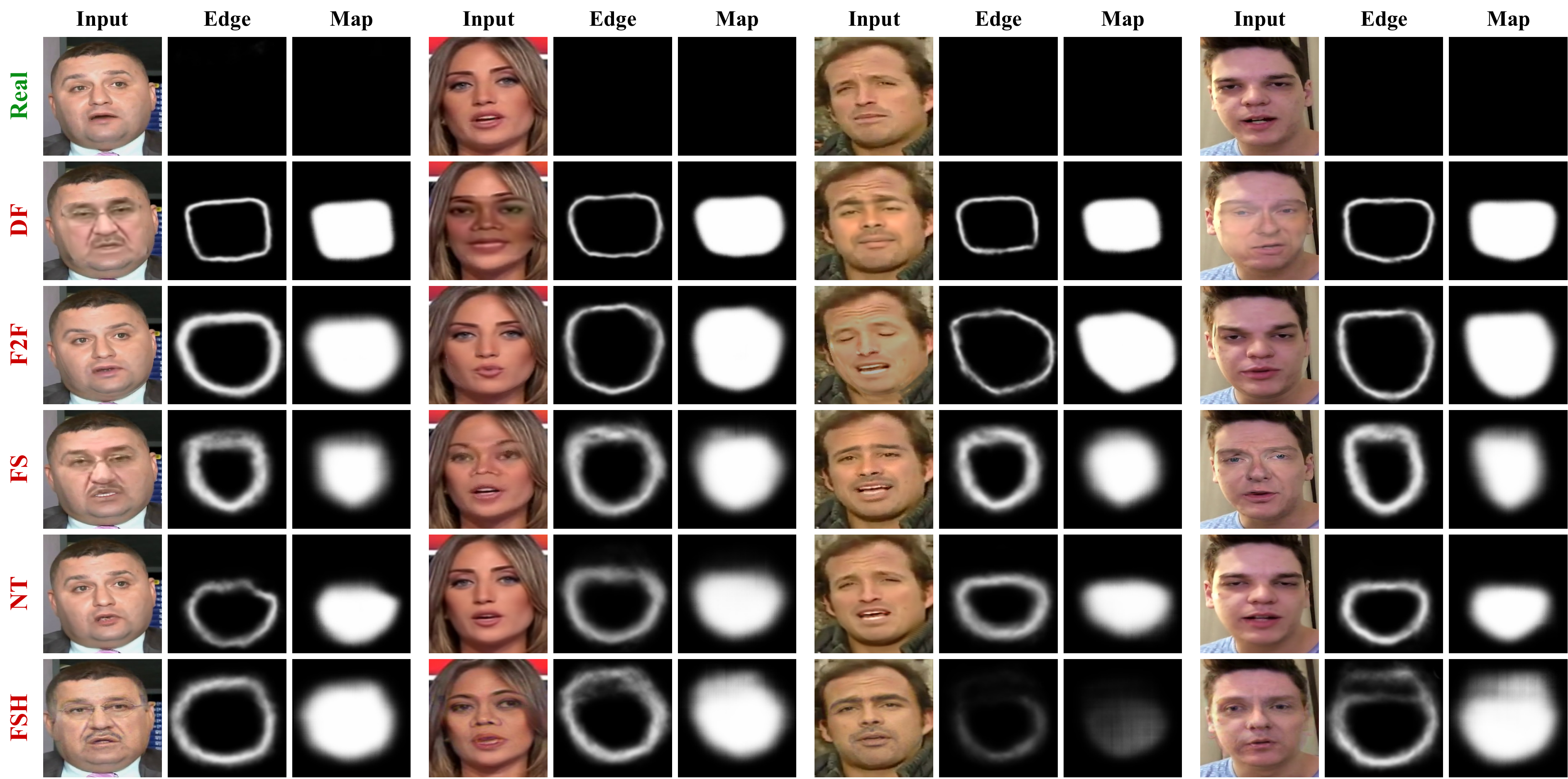}
	\end{center}
	\caption{Additional visualization results of our method.}
	\label{fig:fig7}
\end{figure*}

We have included additional examples in the appendix to provide a more comprehensive demonstration of the effectiveness of our method, as shown in Figure \ref{fig:fig7}. It can be observed that our method is capable of accurately detecting specific manipulation regions. For instance, in the case of Deepfakes, a rectangular mask is employed to guide the replacement of the target face with the source face in the central facial region. Face2Face, conversely, employs RGB tracking to comprehensively capture the whole facial performance for the purpose of expression transfer. FaceShifter exhibits an adaptive capacity by autonomously generating manipulation masks and employing post-processing techniques to mitigate the influence of hair and accessories on the falsified outcomes. 

Our method demonstrates effective localization capabilities across various manipulations. Even in instances of highly convincing manipulation outcomes, such as the FSH in the third column of Figure \ref{fig:fig7}, the model, while expressing a lack of confidence, is still able to provide accurate assessments.

